\documentclass[a4paper, fleqn]{cas-dc}

\usepackage[numbers]{natbib}

\usepackage{glossaries, bm}
\usepackage[caption=false, font=footnotesize]{subfig}

\def\tsc#1{\csdef{#1}{\textsc{\lowercase{#1}}\xspace}}
\tsc{WGM}
\tsc{QE}

\newdefinition{rmk}{Remark}
\newproof{pf}{Proof}
\newproof{pot}{Proof of Theorem \ref{thm}}

\newacronym[longplural={Unmanned Aerial Vehicles}]{UAV}{UAV}{Unmanned Aerial Vehicle}
\newacronym[longplural={Gaussian Mixture Models}]{GMM}{GMM}{Gaussian Mixture Model}
\newacronym[longplural={Probability Density Functions}]{PDF}{PDF}{Probability Density Function}
\newacronym[longplural={Optimal Control Problems}]{OCP}{OCP}{Optimal Control Problem}
\newacronym{FOV}{FOV}{Field of View}
\newacronym{MPC}{MPC}{Model Predictive Control}
\newacronym{ODE}{ODE}{Ordinary Differential Equation}
\newacronym{NLP}{NLP}{Nonlinear Programming}
\newacronym{EKF}{EKF}{Extended Kalman Filter}
\newacronym{PID}{PID}{Proportional-Integral-Derivative}
\newacronym{ROS}{ROS}{Robot Operating System}
\newacronym{GPS}{GPS}{Global Positioning System}

\begin{document}
\let\WriteBookmarks\relax
\def\floatpagepagefraction{1}
\def\textpagefraction{.001}

\shorttitle{}    
\shortauthors{H. Matias et al.}  
\title [mode = title]{Model-Predictive Trajectory Generation\\for Aerial Search and Coverage }  

\tnotemark[1] \tnotetext[1]{This work was partially supported by the Portuguese Fundação para a 
Ciência e a Tecnologia (FCT) through the FirePuma project (DOI: 10.54499/PCIF/MPG/0156/2019), 
through the LARSyS FCT funding (DOI: 10.54499/LA/P/0083/2020, 10.54499/UIDP/50009/2020, and 
10.54499/UIDB/50009/2020), and through the COPELABS, Lusófona University project UIDB/04111/2020.} 

\author[1,2]{Hugo Matias}\cormark[1]\ead{hugomatias@tecnico.ulisboa.pt}
\credit{Writing – original draft, Validation, Software, Methodology, Investigation, Conceptualization}

\author[1,2,3]{Daniel Silvestre}\ead{dsilvestre@isr.tecnico.ulisboa.pt}
\credit{Writing – review and editing, Validation, Supervision, Methodology, Investi
gation, Funding acquisition, Conceptualization}

\affiliation[1]{organization={Institute for Systems and Robotics, LARSyS},
            postcode={1049-001},
            city={Lisbon},
            country={Portugal}}
\affiliation[2]{organization={School of Science and Technology, NOVA University Lisbon},
            postcode={2829-516},
            city={Caparica},
            country={Portugal}}
\affiliation[3]{organization={COPELABS, Lusófona University},
            postcode={1749-024},
            city={Lisbon},
            country={Portugal}}

\cortext[1]{The corresponding author is with the Institute for Systems and Robotics, LARSyS, 
1049-001 Lisbon, Portugal, and with the School of Science and Technology, NOVA University of 
Lisbon, 2829-516 Costa da Caparica, Portugal.}


\begin{abstract}
This paper introduces a trajectory planning algorithm for search and coverage missions with an 
Unmanned Aerial Vehicle (UAV) based on an uncertainty map that represents prior knowledge of the 
target region, modeled by a Gaussian Mixture Model (GMM). The trajectory planning problem is 
formulated as an Optimal Control Problem (OCP), which aims to maximize the uncertainty reduction 
within a specified mission duration. However, this results in an intractable OCP whose objective 
functional cannot be expressed in closed form. To address this, we propose a Model Predictive 
Control (MPC) algorithm based on a relaxed formulation of the objective function to approximate the 
optimal solutions. This relaxation promotes efficient map exploration by penalizing overlaps in the 
UAV’s visibility regions along the trajectory. The algorithm can produce efficient and smooth 
trajectories, and it can be efficiently implemented using standard Nonlinear Programming solvers, 
being suitable for real-time planning. Unlike traditional methods, which often rely on discretizing 
the mission space and using complex mixed-integer formulations, our approach is computationally 
efficient and easier to implement. The MPC algorithm is initially assessed in MATLAB, followed by 
Gazebo simulations and actual experimental tests conducted in an outdoor environment. The results 
demonstrate that the proposed strategy can generate efficient and smooth trajectories for search 
and coverage missions.
\end{abstract}



\maketitle


\section{Introduction}

\glspl{UAV}, commonly known as drones, are an emerging technology with significant potential, 
offering a range of applications across various sectors. These versatile aerial platforms, often 
equipped with high-resolution cameras, sensors, and cutting-edge technology, have the capacity to 
perform operations autonomously, reducing the need for constant human intervention 
\cite{shakhatreh2019unmanned}, \cite{ghamari2022unmanned}. Particularly, drones are significantly 
valuable for search and coverage missions due their ability to cover extensive regions with 
unprecedented ease and speed. This kind of mission finds relevance in numerous applications, 
including search and rescue, wildfire prevention, surveillance, and mapping, among others 
\cite{yao2017optimal}, \cite{afghah2019wildfire}, \cite{nigam2011control}, \cite{lee2021real}, \cite{mansouri2018cooperative}, \cite{mansouri20182d}.

In such a context, the main challenge involves devising trajectories to efficiently cover a 
designated region. This amounts to a complex decision-making and control problem, requiring 
consideration of several factors, including mission objectives, vehicle dynamics, and time 
constraints. Particularly, this paper focus on the coverage planning problem based on a uncertainty 
map describing prior region information, where the goal is to maximize the uncertainty reduction 
within a given flight time.

\subsection{Related Work}

In the literature, several approaches have been proposed to address coverage planning problems, 
which can generally be grouped into two categories \cite{galceran2013survey}. The first category 
comprises exhaustive strategies, where the \gls{UAV} systematically covers the target region. These 
methods are mainly geometric, i.e., the trajectory generation consists of generating geometric 
paths and subsequently parameterizing these paths over time. Common strategies include spiral 
patterns \cite{cabreira2018energy} and back-and-forth movements \cite{torres2016coverage}. 
Additionally, graph-based methods, such as the A* algorithm \cite{acar2006sensor}, have also been 
applied to coverage problems. However, while simple and computationally efficient, these methods 
become evidently inefficient when there are areas of no interest since the entire region is covered 
without any heuristic.

The second group focuses on generating efficient coverage trajectories based on a utility function 
that represents prior knowledge about the target region. These methods typically involve 
discretizing the mission space into a grid, with each grid cell assigned a corresponding importance value. Subsequently, the trajectory 
generation process aims to prioritize the most important areas and is typically based on Bayesian 
like updates. A variety of mathematical and heuristic techniques have been explored, including greedy algorithms \cite{jia2022uav}, probabilistic methods \cite{kim2014response}, and mixed-integer receding horizon approaches \cite{yao2017gaussian}. However, a key drawback of grid-based approaches is their sensitivity to localization errors, which can significantly affect accuracy. Moreover, the grid representation of the environment demands large amounts of memory. Additionally, when mixed-integer formulations are considered, similar to \cite{yao2017gaussian}, the computational burden becomes even more significant, as the computational complexity becomes exponential with respect to the number of mixed-integer constraints. This makes it increasingly difficult to generate trajectories in a reasonable time frame, limiting the applicability of these methods for real-time trajectory planning. Therefore, there is still no standard for an optimization-based solution to the problem. 

\subsection{Paper Overview}

This paper introduces a trajectory planning algorithm for search and coverage missions with a 
\gls{UAV} based on an uncertainty map that represents prior knowledge of the target region, modeled 
by a \gls{GMM}. The trajectory planning problem is framed as an \gls{OCP}, which aims to maximize 
the uncertainty reduction within a given mission duration. However, this results in an intractable 
\gls{OCP} whose objective functional cannot be expressed in closed form. To address this, we 
propose a \gls{MPC} algorithm based on a relaxed formulation of the objective function to 
approximate the optimal solutions. This relaxation promotes efficient map exploration by penalizing 
overlaps in the \gls{UAV}’s visibility regions along the trajectory. The algorithm is able to 
produce efficient and smooth trajectories, and it can be efficiently implemented using standard 
\gls{NLP} solvers, being suitable for real-time planning. Unlike traditional methods, which often 
rely on discretizing the mission space and using mixed-integer formulations, our approach is 
computationally efficient and easier to implement.

The remainder of this paper is organized as follows. Section \ref{Sec:ProblemFormulation} 
introduces important preliminaries and formulates the trajectory planning problem from an optimal 
control standpoint. Section \ref{Sec:ProposedSolution} outlines the proposed \gls{MPC} approach, 
and Section \ref{Sec:ControlArchitecture} details the control architecture implemented to execute 
the \gls{MPC} algorithm on a quadrotor. In Section \ref{Sec:SimulationResults}, we present a series 
of simulation results obtained in MATLAB, followed by Section \ref{Sec:ExperimentalValidation}, 
which showcases additional results from both Gazebo simulations and actual outdoor experiments. 
Finally, Section \ref{Sec:Conclusion} summarizes conclusions and suggests directions for future 
research.

\subsection{Notation}

$\mathbb{Z}$ is the set of all integers and $\mathbb{Z}_{[i, j]}$ is the set of integers from $i$ 
to $j$. $\mathbb{R}$, $\mathbb{R}_{\geq 0}$, and $\mathbb{R}_{> 0}$ are the sets of real, 
nonnegative, and positive numbers, respectively. $\mathbb{R}^n$ is the $n$-dimensional euclidean 
space, and $\mathbb{S}^{n-1}$ is the unit sphere in $\mathbb{R}^n$. $\mathbb{R}^{n\times m}$ is the 
set of $n\times m$ real matrices, $\mathbb{R}^{n\times n}_{\succ 0}$ is the set of 
positive-definite square matrices of size $n$, and $\text{SO}(n)$ denotes the special orthogonal 
group in $\mathbb{R}^n$. For a set $\mathcal{S} \subseteq \mathbb{R}^n$, $\text{int}(\mathcal{S})$ 
and $\partial\mathcal{S}$ are the interior and boundary of $\mathcal{S}$, respectively. The 
$p$-norm of a vector $\mathbf{x} \in \mathbb{R}^n$ is denoted as $\|\mathbf{x}\|_p$ 
($\|\mathbf{x}\| = \|\mathbf{x}\|_2)$, and for two column vectors 
$\mathbf{x}_1 \in\mathbb{R}^{n_1}$, $\mathbf{x}_2 \in \mathbb{R}^{n_2}$, we often use the notation 
$(\mathbf{x}_1, \mathbf{x}_2) = [\mathbf{x}_1^\top\, \mathbf{x}_2^\top]^\top \in 
\mathbb{R}^{n_1+n_2}$. Finally, $\mathbf{0}_{n\times m}$ is the $n\times m$ zero matrix, and 
$\mathbf{I}_{n}$ is the identity matrix of size $n$ (dimensions may be omitted when clear from 
context).

\newpage


\section{Preliminaries and Problem Formulation} \label{Sec:ProblemFormulation}

This section formalizes the trajectory planning problem addressed in this paper. It begins by 
establishing assumptions concerning the uncertainty map and the sensing model of the \gls{UAV}. 
Subsequently, we formulate the trajectory planning problem from an optimal control standpoint.


\subsection{Uncertainty Map}

\begin{figure}[h] 
    \centering 
    \subfloat[Graph]{
        \includegraphics[width=0.49\linewidth]{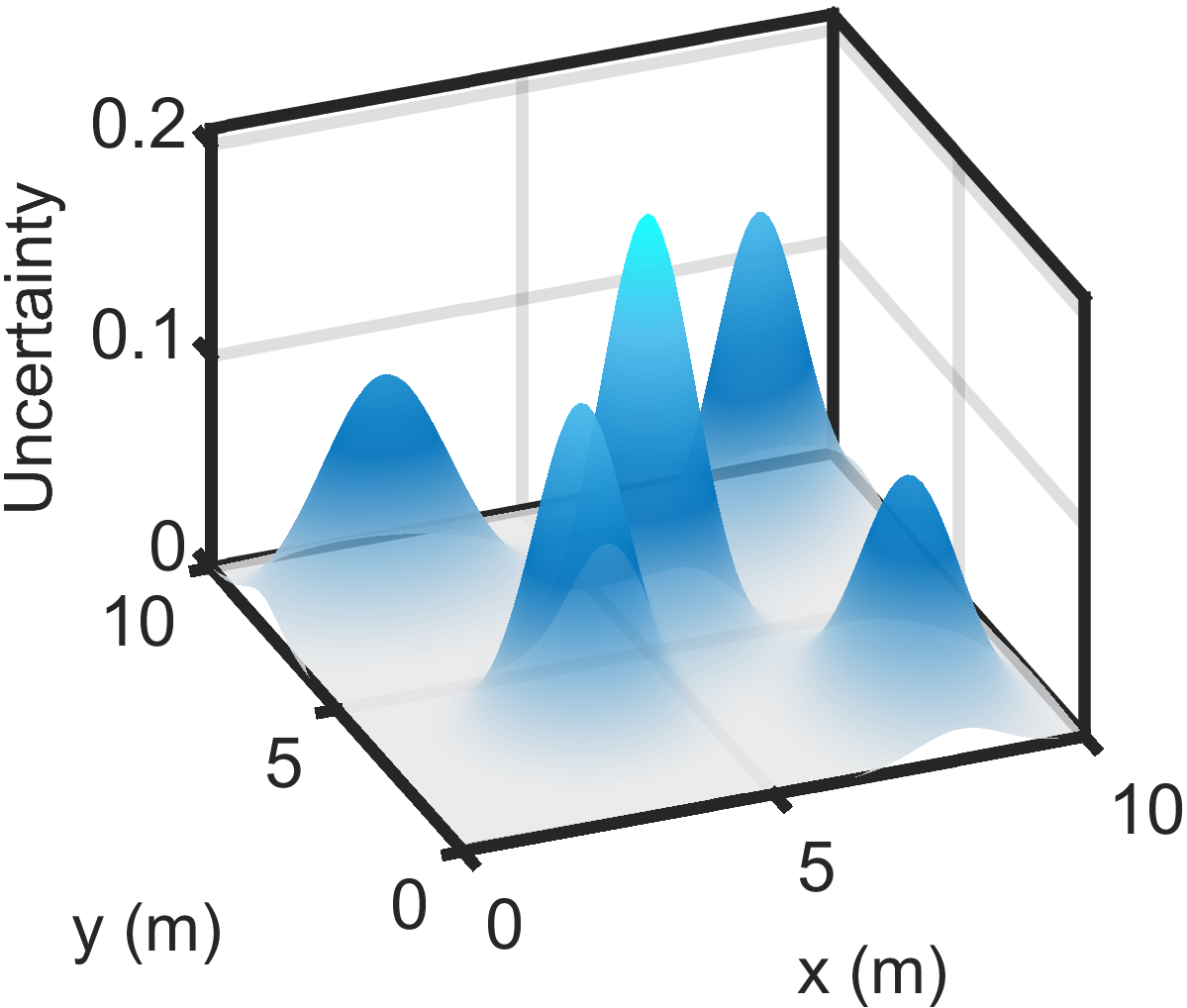}}
    \subfloat[Level curves]{
        \includegraphics[width=0.47\linewidth]{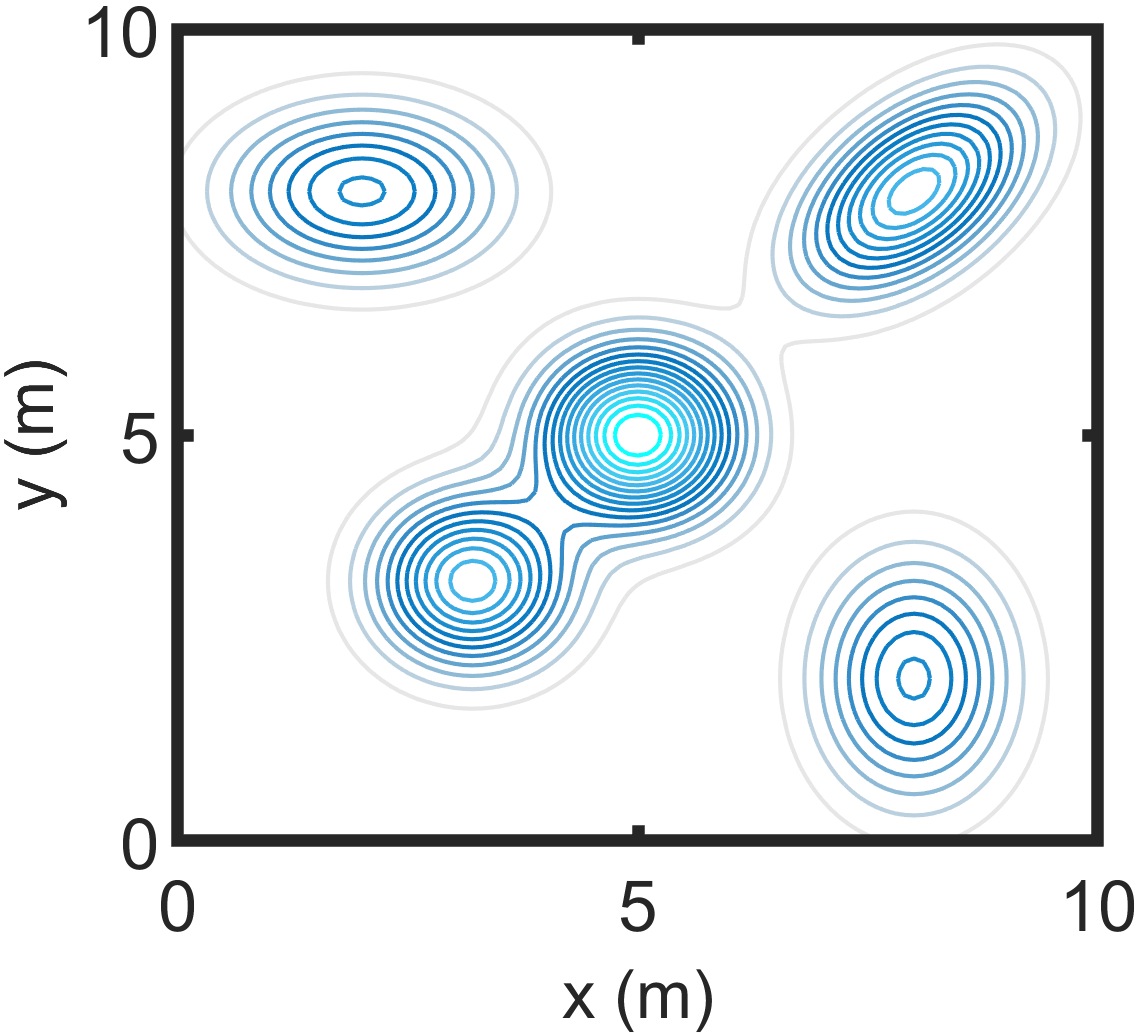}}
    \caption{Example of an uncertainty map.} 
    \label{Fig:UncertaintyMap} 
\end{figure}

The uncertainty map is a function $h: \mathbb{R}^2 \rightarrow \mathbb{R}_{\geq 0}$ that describes 
the prior significance of each point $\mathbf{p} \in \mathbb{R}^2$ to be analyzed by the vehicle. 
Since the original structure of the uncertainty map typically may not follow common and well-known 
models, we consider that the uncertainty map can be arbitrarily well approximated by a \gls{GMM}. 
Specifically, for a model with $M$ components, $h$ is defined by
\begin{equation}
    h(\mathbf{p}) = \sum_{i=1}^{M} w_i\, \mathcal{N}(\mathbf{p}; \bm{\mu}_i, \bm{\Sigma}_i)
\end{equation}
for all $\mathbf{p} \in \mathbb{R}^2$, where $\mathcal{N}$ denotes a standard two-dimensional 
Gaussian distribution. The parameters $w_i \in \mathbb{R}_{> 0}$, $\bm{\mu}_i \in \mathbb{R}^2$, 
and $\mathbf{\Sigma}_i \in \mathbb{R}^{2\times2}_{\succ 0}$ are, respectively, the weight, the mean 
vector, and the covariance matrix of the $i\text{th}$ Gaussian component. Also, we highlight that 
the uncertainty map is not required to be a \gls{PDF}. However, for convenience, we assume that $h$ 
is normalized, meaning that the prior uncertainty volume is one and thus $\sum_{i=1}^M w_i = 1$. 
Fig. \ref{Fig:UncertaintyMap} illustrates a plausible instance of an uncertainty map.


\subsection{Sensing Model}

\begin{figure}[h] 
    \centering 
    \includegraphics[width=0.5\linewidth]{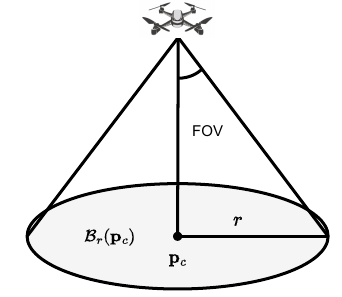} 
    \caption{Sensor \gls{FOV} and visibility region.} 
    \label{Fig:DroneFOV}
\end{figure}

This work assumes that the drone flies at a constant altitude and features a gimbal camera, which 
remains directed downwards even when the vehicle is performing pitch or roll maneuvers. Moreover, 
at each time instant $t$, we assume the camera covers a circular region, 
$\mathcal{B}_r(\mathbf{p}_c)$, defined by
\begin{equation}
    \mathcal{B}_r(\mathbf{p}_c) = \left\{ \mathbf{p} \in \mathbb{R}^2: 
    \left\| \mathbf{p} - \mathbf{p}_c \right\| < r \right\},
\end{equation}
where $\mathbf{p}_c \in \mathbb{R}^2$ is the vehicle's horizontal position and $r$ is the 
observation radius, as displayed in Fig. \ref{Fig:DroneFOV}. In addition, the vehicle is assumed to 
have a perfect quality of exploration, meaning that after it analyzes a given area, the uncertainty 
reduces to zero for all points inside the observation region. 

\subsection{Optimal Control Problem}

The problem addressed in this letter amounts to generating optimal coverage trajectories for the 
\gls{UAV}. The trajectories should maximize an objective functional regarding the mission objective 
while satisfying constraints accounting for their dynamic feasibility. Consequently, this problem 
can be formulated as the following \gls{OCP}:
\begin{equation} 
    \begin{aligned}
        \underset{\mathbf{x}, \mathbf{u}}{\text{maximize}} 
        \quad & J(\mathbf{x}, \mathbf{u})\\
        \text{subject to} \quad & \mathbf{x}(0) = \mathbf{x}_0,\\
        & \Dot{\mathbf{x}}(t) = \mathbf{F}(\mathbf{x}(t), \mathbf{u}(t)),\,\, 
        \forall t \in [0,T],\\
        & \mathbf{x}(t) \in \mathcal{X},\,\, \forall t \in [0,T],\\
        & \mathbf{u}(t) \in \mathcal{U},\,\, \forall t \in [0,T],
    \end{aligned}
    \label{Eq:ProblemFormulation}
\end{equation}
where $T$ denotes the total flight duration, $\mathbf{x}: [0,T] \rightarrow \mathbb{R}^{n_x}$ and 
$\mathbf{u}: [0,T] \rightarrow \mathbb{R}^{n_u}$ designate the state and input signals of the 
vehicle's model, $\mathbf{x}_0$ is the initial state, and the sets $\mathcal{X}$ and $\mathcal{U}$ 
constitute the admissible states and inputs for the vehicle. 

\begin{figure}[h] 
    \centering 
    \includegraphics[width=\linewidth]{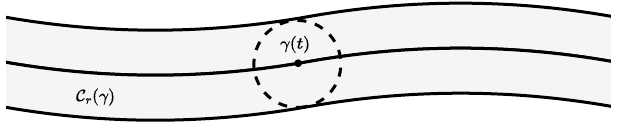} 
    \caption{Illustration of the set $\mathcal{C}_r(\bm{\gamma})$.} 
    \label{Fig:CirclesUnion}
\end{figure}

Let now $\bm{\gamma}: [0,T] \rightarrow \mathbb{R}^{2}$ denote the vehicle's trajectory, related to 
the state via an auxiliary matrix $\mathbf{C}_{\bm{\gamma}} \in \mathbb{R}^{2 \times n_x}$ as
\begin{equation}
    \bm{\gamma}(t) = \mathbf{C}_{\bm{\gamma}} \mathbf{x}(t)
\end{equation}
for all $t \in [0,T]$. The goal is to maximize the uncertainty reduction, i.e., the difference 
between the uncertainty volume in the map before and after the mission. Thus, $J$ is given by
\begin{equation}
    J(\bm{\gamma}) = \int_{\mathcal{C}_r(\bm{\gamma})} h(\mathbf{p})\,d\mathbf{p},
    \label{Eq:ObjectiveFunctional}
\end{equation}
where the set $\mathcal{C}_r(\bm{\gamma})$ is defined as the union of all observation regions along 
the trajectory of the vehicle, i.e., 
\begin{equation}
    \mathcal{C}_r(\bm{\gamma}) = \bigcup_{t = 0}^{T}\mathcal{B}_r(\bm{\gamma}(t)),
\end{equation}
as illustrated in Fig. \ref{Fig:CirclesUnion}.
\vfill

The optimal control problem in \eqref{Eq:ProblemFormulation} is particularly difficult to solve 
because the objective functional, as defined in \eqref{Eq:ObjectiveFunctional}, does not have a 
closed-form expression. Thus, we need to consider a relaxed formulation. Additionally, in order to 
make the problem computationally tractable, it needs to be discretized as well. However, even with 
the relaxation and discretization, solving the problem globally for a large flight time is 
computationally challenging. As a result, we adopt a local approach based on \gls{MPC} to 
approximate the solutions of \eqref{Eq:ProblemFormulation} while adding the possibility for online 
execution.


\section{MPC Approach with Relaxed Formulation} \label{Sec:ProposedSolution}

To tackle the problem defined in the previous section, we consider an \gls{MPC} approach with a 
relaxed formulation of the objective function. \gls{MPC} consists of solving a discrete-time 
\gls{OCP} at each sampling time. Each optimization results in a sequence of future optimal control 
actions and a corresponding sequence of future states. Only the first sample from the predicted 
optimal control sequence is applied to the vehicle, and then the process repeats at the next 
sampling time.

More specifically, at each discrete-time instant $k$, for a given initial state $\mathbf{x}[k]$ of 
the vehicle, the control policy is defined by solving an optimization problem of the form
\begin{align}
    \underset{\hat{\mathbf{x}}_k, \hat{\mathbf{u}}_k}{\text{maximize}} 
    \quad & J_k(\hat{\mathbf{x}}_k, \hat{\mathbf{u}}_k)\label{Eq:MPCFormulation}\\
    \text{subject to} \quad & \hat{\mathbf{x}}_k[0] = \mathbf{x}[k],\nonumber\\
    & \hat{\mathbf{x}}_k[n+1] = 
    \mathbf{f}(\hat{\mathbf{x}}_k[n], \hat{\mathbf{u}}_k[n]),\,\, 
    \scalebox{0.8}{$\forall n \in \mathbb{Z}_{[0, N-1]}$},\nonumber\\
    & \hat{\mathbf{x}}_k[n] \in \mathcal{X},\,\, \forall n \in \mathbb{Z}_{[0, N]},\nonumber\\
    & \hat{\mathbf{u}}_k[n] \in \mathcal{U},\,\, \forall n \in \mathbb{Z}_{[0, N-1]},\nonumber
\end{align}
where $N$ is the horizon length, 
$\hat{\mathbf{x}}_k: \mathbb{Z}_{[0, N]} \rightarrow \mathbb{R}^{n_x}$ and 
$\hat{\mathbf{u}}_k: \mathbb{Z}_{[0, N-1]} \rightarrow \mathbb{R}^{n_u}$ are the predicted 
state and control sequences at the time step $k$, and the function $\mathbf{f}$ describes a 
discrete-time model of the vehicle dynamics. Moreover, the sets $\mathcal{X}$ and $\mathcal{U}$ 
constitute the admissible states and inputs for the vehicle, as defined in 
\eqref{Eq:ProblemFormulation}. The input applied to the vehicle at the discrete-time instant $k$, 
$\mathbf{u}[k]$, is given by
\begin{equation}
    \mathbf{u}[k] = \hat{\mathbf{u}}_k^*[0],
\end{equation}
where $\hat{\mathbf{u}}_k^*[0]$ is the first sample of the predicted optimal control sequence. In a 
general sense, the optimization problem in \eqref{Eq:MPCFormulation} is a structured \gls{NLP}, 
which may be solved efficiently using commercially available \gls{NLP} solvers.


\subsection{Objective Function}

Our approach for approximating the problem described in Section \ref{Sec:ProblemFormulation} relies 
on a relaxed formulation of the objective function. More specifically, the objective function is 
defined by the combination of two objectives as
\begin{equation}
    J_k(\hat{\bm{\gamma}}_k) = \Tilde{J}_k(\hat{\bm{\gamma}}_k) - \lambda 
    P_k(\hat{\bm{\gamma}}_k),
    \label{Equation:ObjectiveFunction}
\end{equation}
where $\hat{\bm{\gamma}}_k: \mathbb{Z}_{[0, N]} \rightarrow \mathbb{R}^2$ denotes the predicted 
sequence of vehicle positions at the discrete-time instant $k$, and 
$\lambda \in \mathbb{R}_{\geq 0}$ is a scaling factor that weighs the relative importance of the 
two objectives.

The first term in \eqref{Equation:ObjectiveFunction}, $\Tilde{J}_k$, expresses the objective of 
prioritizing the regions with the highest uncertainty. Namely, it is determined by summing the 
uncertainty volumes that are predicted to be covered by the vehicle at each time step of the 
prediction horizon, i.e., 
\begin{equation}
    \Tilde{J}_k(\hat{\bm{\gamma}}_k) = 
    \sum_{n=0}^{N}\int_{\mathcal{B}_r(\hat{\bm{\gamma}}_k[n])} h(\mathbf{p})\,d\mathbf{p}.
    \label{Equation:FirstTerm}
\end{equation}
However, this term does not consider the previously covered regions nor the intersections between 
the observation regions within the prediction horizon. Therefore, if the objective function was 
defined solely by this term, the trajectories would converge to a point where the uncertainty 
volume covered by the vehicle is locally maximum and remain there.

\begin{figure}[h] 
    \centering 
    \includegraphics[width=\linewidth]{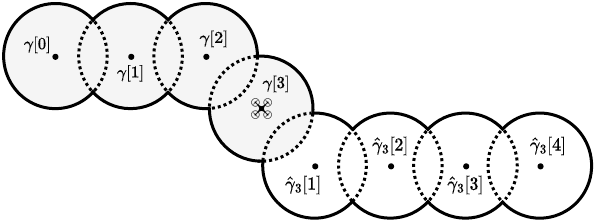} 
    \caption{Illustration of the observation regions at the discrete-time instant $k = 3$ for a 
    horizon length $N = 4$ (grey - previously covered circles; white - predicted observation 
    circles).} 
    \label{Fig:HorizonCircles}
\end{figure}

To encode the information about the previously explored regions along with the intersections 
between the observation areas within the prediction horizon, we add a penalty term $P_k$ to 
the \gls{MPC} objective function. This term penalizes intersections between all possible pairs of 
observation circles along the vehicle's trajectory. Thus, two kinds of intersections can be 
distinguished: intersections between the predicted observation circles and previously covered ones, 
and intersections between the observation circles over the prediction horizon. Hence, the 
penalty term can be written as
\begin{equation}
    P_k(\hat{\bm{\gamma}}_k) = P_k^B(\hat{\bm{\gamma}}_k) + P_k^H(\hat{\bm{\gamma}}_k),
    \label{Equation:PenaltyTerm}
\end{equation}
where $P_k^B$ penalizes intersections between the predicted observation regions and the previously 
covered ones, and $P_k^H$ penalizes intersections within the prediction horizon. Therefore, 
assuming that $p: \mathbb{R}^2 \times \mathbb{R}^2 \rightarrow \mathbb{R}_{\geq 0}$ is a function 
that penalizes the intersection between two circles and $\bm{\gamma}[i]$ is the vehicle's position 
at the time step $i$, $P_k^B$ is defined as
\begin{equation}
    P_k^B(\hat{\bm{\gamma}}_k) = \sum_{n=1}^{N}\sum_{i=0}^{k} 
    p(\hat{\bm{\gamma}}_k[n], \bm{\gamma}[i]),
\end{equation}
and $P_k^H$ is given by
\begin{equation}
    P_k^H(\hat{\bm{\gamma}}_k) = \sum_{n=2}^{N}\sum_{i=1}^{n-1}
    p(\hat{\bm{\gamma}}_k[n], \hat{\bm{\gamma}}_k[i]).
\end{equation}
The proposed approach is illustrated in Fig. \ref{Fig:HorizonCircles}, which represents the 
previously covered regions and the ones predicted to be covered by the vehicle at a given iteration 
of the \gls{MPC} algorithm. It now remains to design the penalty function $p$.

Before proceeding, it is worth highlighting that the integral evaluations in 
\eqref{Equation:FirstTerm} still cannot be expressed in closed form. Nevertheless, as the
integrals are now computed over circular domains, it becomes possible to approximate them through 
numerical methods such as quadrature rules or by discretizing the observation region using a grid. 
However, we will typically focus on scenarios where the observation radius is small compared 
to the structure of the uncertainty map and, consequently, \eqref{Equation:FirstTerm} can be 
approximated as
\begin{equation}
    \Tilde{J}_k(\hat{\bm{\gamma}}_k) \simeq \pi r^2 \sum_{n=0}^{N}h(\hat{\bm{\gamma}}_k[n]).
\end{equation}

\subsection{Penalty Function}

A plausible way to design the penalty function $p$ could be to define it as the overlap area 
between two circles. Let $\mathbf{c}_1, \mathbf{c}_2 \in \mathbb{R}^2$ be the centers of two 
circles, both with radius $r$, and $d = \left\| \mathbf{c}_1 - \mathbf{c}_2\right\|$ the distance 
between them. The overlap area between the circles, 
$a: \mathbb{R}_{\geq 0} \rightarrow \mathbb{R}_{\geq 0}$, is given by
\begin{equation}
    \scalebox{0.98}{$
    a(d) = 
    \begin{cases}
        2r^2\arccos\left(\frac{d}{2r}\right) - d\sqrt{r^2-d^2},\,\, 
        \text{if}\,\, d \leq 2r,\\
        0,\,\, \text{if}\,\,  d > 2r.
    \end{cases}
    $}
    \label{Eq:OverlapArea}
\end{equation}
However, an expression of this complexity would represent a computational bottleneck. Additionally, 
since the function in \eqref{Eq:OverlapArea} is defined piecewise, the logic condition would need 
to be converted into a constraint using auxiliary binary variables. This would lead to a 
mixed-integer \gls{MPC} formulation, significantly increasing the computational load, as in 
\cite{yao2017gaussian}.

\begin{figure}[h] 
    \centering 
    \includegraphics[width=\linewidth]{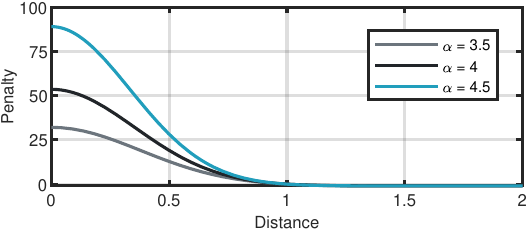} 
    \caption{Evolution of the penalty function with the distance between the two circles
    for some values of $\alpha$ with $r = 0.5$ m.} 
    \label{Fig:PenaltyFunction}  
\end{figure}

Nevertheless, it is not necessary to compute the overlap area between two circles to penalize the 
intersection between them. Instead, the penalization can be achieved with a function that directly 
penalizes the intersection. To this end, we formulate the penalty function by imposing an 
exponential penalty on the violation of the condition 
$\left\| \mathbf{c}_1 - \mathbf{c}_2\right\| > 2r$. Specifically, the penalty function is defined as
\begin{equation}
    p(\mathbf{c}_1, \mathbf{c}_2) = \exp{\left\{\alpha\left((2r)^2-
    \left\| \mathbf{c}_1 - \mathbf{c}_2\right\|^2\right)\right\}} - 1,
\end{equation}
where $\alpha > 0$ is a parameter that can be tuned. Additionally, the subtraction of 1 is included 
so that the penalty function has a value of zero when 
$\left\| \mathbf{c}_1 - \mathbf{c}_2\right\| = 2r$, but it has no effect on
the optimization since it is a constant. Fig. \ref{Fig:PenaltyFunction} illustrates the 
evolution of the penalty function with the distance between the two circles for some values of 
$\alpha$.
\vfill


\subsection{Computational Complexity}

From a computational standpoint, it is important to analyze the complexity of the proposed 
algorithm. Besides the inherent complexity of the problem, determined by the structure of the 
uncertainty map and the imposed constraints, it is relevant to examine the number of terms 
comprising the \gls{MPC} objective function, which directly influences the number of evaluations 
that the solver must perform. In particular, it is worth noting that the number of terms comprising 
$\Tilde{J}_k$ and $P_k^H$ is determined by the horizon length. Specifically, the number of terms in 
$\Tilde{J}_k$ increases linearly with the horizon length, whereas $P_k^H$ consists of $N(N-1)/2$ 
terms, leading to a quadratic growth with respect to the horizon length. 

Besides the quadratic growth of $P_k^H$ with the horizon, a significant computational burden arises 
from $P_k^B$. At each time instant $k$, the number of terms comprising $P_k^B$ increases by $N$, 
meaning that $P_k^B$ grows linearly with the flight time assigned for the surveillance mission. A 
direct solution is to set a maximum backward horizon length, $N_B$, limiting $P_k^B$ to a given 
number of terms. This consists in defining $P_k^B$ as 
\begin{equation}
    P_k^B(\hat{\bm{\gamma}}_k) = \sum_{n=1}^{N}\sum_{i=k-N_B+1}^{k} 
    p(\hat{\bm{\gamma}}_k[n], \bm{\gamma}[i]).
\end{equation}
Nonetheless, if the backward horizon is not long enough, the vehicle may revisit previously 
explored regions. Hence, a more effective approach to be considered in future research involves 
developing a subroutine that progressively reduces the number of components in the penalty term 
while preserving information about all previously covered regions.

Additionally, it is important to clarify that despite the notion that the number of terms in the 
objective function grows at each time step, the optimization solvers are built by allocating the 
necessary resources for the entire mission duration. This decision follows from the substantial 
additional overhead that there would be in generating a solver at each time step. Hence, the number 
of terms in the objective function actually remains constant throughout the whole mission, with the 
terms related to future time steps in $P_k^B$ being assigned a null weight. As a result, despite 
potential fluctuations introduced by the problem, the computation times are expected to remain 
approximately constant.

\subsection{Evaluation Metric}

It is essential to establish an overall metric to evaluate the performance of the algorithm and 
perform comparisons. In this context, a reliable approach for assessing the quality of the 
generated trajectories involves computing the time evolution of the uncertainty volume covered 
by the vehicle. By disregarding the uncertainty coverage between sampling instants, this metric can 
be approximated as
\begin{equation}
    H_{\bm{\gamma}}[k] = \int_{\bigcup_{i = 0}^{k}\mathcal{B}_r(\bm{\gamma}[i])} 
    h(\mathbf{p})\,d\mathbf{p}.
    \label{Equation:UncertaintyReduction}
\end{equation}
Furthermore, since there is no closed-form expression for \eqref{Equation:UncertaintyReduction}, 
the metric is numerically approximated by discretizing the map into a grid.


\section{Quadrotor Motion Control} \label{Sec:ControlArchitecture}

This work focuses on multirotor aerial vehicles due to their agility and hovering capabilities. 
Moreover, a quadrotor is available to perform experimental tests. This section outlines the control 
architecture employed to implement the proposed algorithm on a quadrotor aerial vehicle.

\begin{figure}[h] 
    \centering 
    \includegraphics[width=\linewidth]{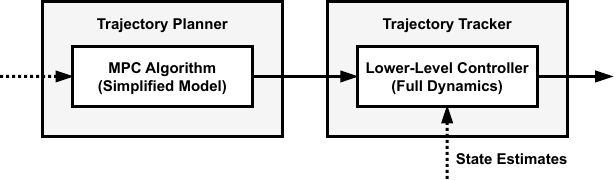} 
    \caption{Full motion control scheme of the \gls{UAV}.}
    \label{Fig:ControlArchitecture}  
\end{figure}

We consider a dual-layer structure of motion control, as illustrated in Fig. 
\ref{Fig:ControlArchitecture}. The proposed \gls{MPC} algorithm serves as a higher-level controller 
(trajectory planner), which generates high-level references for the \gls{UAV}. The lower-level 
controller (trajectory tracker) directly applies control inputs to the vehicle to track the 
references provided by the upper-level controller. For the purpose of efficiency, the \gls{MPC} 
algorithm considers a simplified model of the vehicle, while the lower-level controller accounts 
for the full quadrotor dynamics.

\subsection{Full Dynamics Model}

\begin{figure}[h] 
    \centering 
    \includegraphics[width=0.75\linewidth]{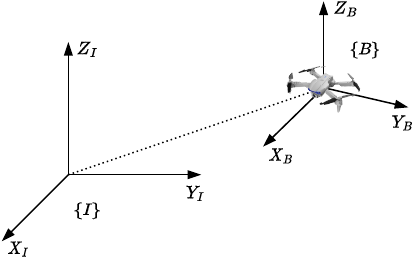} 
    \caption{Quadrotor reference frames.} 
    \label{Fig:ReferenceFrames}  
\end{figure}

For completeness, we begin by describing the full nonlinear dynamics of a quadrotor. The nonlinear 
quadrotor dynamics are described in the body $\{B\}$ and inertial $\{I\}$ frames depicted in Fig. 
\ref{Fig:ReferenceFrames} while assuming that the origin of $\{B\}$ is coincident with the 
quadrotor's center of mass. Let $\bm{\xi} = [\bm{\gamma}\,\, z]^\top$ denote the quadrotor's 
position in the inertial frame and $\bm{\eta} = [\phi\,\, \theta\,\, \psi]^\top$ describe the 
orientation of the body frame with respect to the inertial frame, where $\phi$, $\theta$ and 
$\psi$are the roll, pitch and yaw angles. Moreover, let $\bm{\omega} \in \mathbb{R}^{3}$ denote the 
angular velocity of $\{B\}$ with respect to $\{I\}$, expressed in $\{B\}$. Additionally, let 
$\mathbf{R}(\bm{\eta}) \in \text{SO}(3)$ be the rotation matrix from $\{B\}$ to $\{I\}$ and 
$\mathbf{T}(\bm{\eta}) \in \mathbb{R}^{3\times3}$ a matrix that converts the angular velocity to 
angle rates. Finally, let $m$ be the mass of the vehicle, $\mathbf{I} \in \mathbb{R}^{3\times3}$ 
the inertia matrix expressed in $\{B\}$, and $g$ the gravitational acceleration. Based on the 
Newton-Euler formalism \cite{mahony2012multirotor}, the quadrotor motion is governed by
\begin{equation} 
    \begin{aligned}
        m\Ddot{\bm{\xi}} &= -mg\mathbf{e}_3 + \mathbf{R}(\bm{\eta})\,F\mathbf{e}_3,\\
        \Dot{\bm{\eta}} &= \mathbf{T}(\bm{\eta})\,\bm{\omega},\\
        \mathbf{I}\Dot{\bm{\omega}} &= -\bm{\omega}\times\mathbf{I}\bm{\omega} + \bm{\tau},
    \end{aligned}
\end{equation}
where $F$ is the thrust magnitude and $\bm{\tau} \in \mathbb{R}^{3}$ is the torque applied to the 
\gls{UAV}, described in $\{B\}$. Ultimately, the relation between the rotation speeds of the 
rotors, $\Bar{\omega}_i$, $i = 1, \dots, 4$, and the thrust and torque vector can be modeled as
\begin{equation}
    \begin{bmatrix}
        F\\
        \bm{\tau}
    \end{bmatrix}
    = \bm{\Gamma}
    \begin{bmatrix}
        \Bar{\omega}_1^2 & \Bar{\omega}_2^2 & \Bar{\omega}_3^2 & \Bar{\omega}_4^2
    \end{bmatrix}^\top,
\end{equation}
where $\bm{\Gamma} \in \mathbb{R}^{4\times4}$ depends on the rotors arrangement.

\subsection{Simplified Model}

At the planning level, considering that the \gls{UAV} flies at a constant altitude and ignoring the 
fast rotational dynamics of the vehicle, the \gls{UAV} might be modeled as a two-dimensional 
point-mass system with double-integrator dynamics. Thus, the state vector is composed of the 
positions and velocities on the horizontal plane, 
$\mathbf{x} = [\bm{\gamma}^\top\,\,\Dot{\bm{\gamma}}^\top]^\top$, and the control input is the 
acceleration on the horizontal plane. Consequently, at the planning level, the quadrotor dynamics 
take the form
\begin{equation}
    \begin{bmatrix}
        \Dot{\bm{\gamma}}\\
        \Ddot{\bm{\gamma}}
    \end{bmatrix}
    =
    \begin{bmatrix}
        \mathbf{0}_{2\times2} & \mathbf{I}_2\\
        \mathbf{0}_{2\times2} & \mathbf{0}_{2\times2}
    \end{bmatrix}
    \begin{bmatrix}
        \bm{\gamma}\\
        \Dot{\bm{\gamma}}
    \end{bmatrix}
    +
    \begin{bmatrix}
        \mathbf{0}_{2\times2}\\
        \mathbf{I}_2
    \end{bmatrix}
    \mathbf{u}.
\end{equation} 
This mismatch is not critical for obtaining good performance as long as the generated trajectories 
are not extremely aggressive so that the inner-loop dynamics become visible.

\subsection{Implementation Details}

\begin{figure}[h] 
    \centering 
    \includegraphics[width=\linewidth]{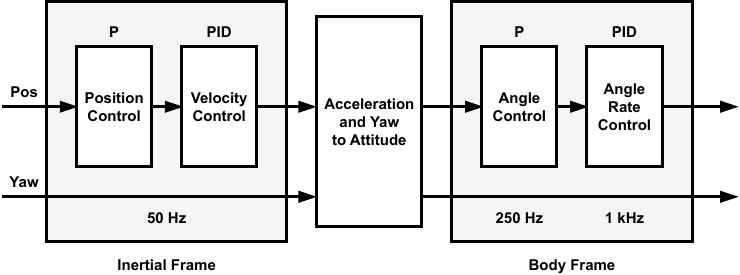} 
    \caption{PX4 controller architecture.} 
    \label{Fig:PX4Architecture}  
\end{figure}

In practical terms, the proposed motion control scheme is implemented using a PX4 Autopilot
\cite{px4}. The PX4 Autopilot provides the lower-level controller and an \gls{EKF} to 
process sensor measurements and provide state estimates to the controllers. As detailed in Fig. 
\ref{Fig:PX4Architecture}, the controller supplied by the PX4 Autopilot follows a standard 
cascaded architecture with several stages. Each stage is composed of a proportional or \gls{PID} 
controller that generates references for the upcoming stage based on references provided by the 
previous stage. From a general perspective, the PX4 controller consists of two main control loops: 
position and attitude. The position control loop commands accelerations, which are then converted 
into attitude and net thrust references. The attitude control loop receives attitude and net thrust 
references and commands thrust references for the vehicle motors.
\vspace{1mm}


\section{Simulation Results} \label{Sec:SimulationResults}

This section assesses the efficacy of the proposed \gls{MPC} algorithm through different simulation 
examples obtained in a MATLAB environment. The goal is to perform an initial analysis of the 
proposed \gls{MPC} algorithm. Therefore, the simulations presented in this section are performed 
assuming that the \gls{UAV} follows ideal double-integrator dynamics with constraints on the 
maximum velocity and acceleration magnitudes. At each discrete-time instant $k$, the \gls{MPC} 
algorithm involves solving the following optimization problem
\begin{equation} 
    \scalebox{0.92}{$
    \begin{aligned}
        \underset{\hat{\mathbf{x}}_k, \hat{\mathbf{u}}_k}{\text{maximize}} 
        \quad & J_k(\hat{\mathbf{x}}_k, \hat{\mathbf{u}}_k)\\
        \text{subject to} \quad & \hat{\mathbf{x}}_k[0] = \mathbf{x}[k],\\
        & \hat{\mathbf{x}}_k[n+1] = 
        \mathbf{A}\hat{\mathbf{x}}_k[n] + \mathbf{B}\hat{\mathbf{u}}_k[n],\,\, 
        \forall n \in \mathbb{Z}_{[0, N-1]},\\
        & \left\| \mathbf{C}_{\Dot{\bm{\gamma}}}\hat{\mathbf{x}}_k[n] 
        \right\| \leq v_{\text{max}},\,\, 
        \forall n \in \mathbb{Z}_{[0, N]},\\
        & \left\| \hat{\mathbf{u}}_k[n] \right\| \leq a_{\text{max}},\,\, 
        \forall n \in \mathbb{Z}_{[0, N-1]},
    \end{aligned}
    $}
\end{equation}
where the objective function is obtained as detailed in Section \ref{Sec:ProposedSolution}. 
The matrices $\mathbf{A}$ and $\mathbf{B}$ correspond to the discrete double-integrator dynamics 
(zero-order hold) and are
\begin{equation}
    \begin{aligned}
        \mathbf{A} = 
        \begin{bmatrix}
            \mathbf{I}_2 & T_s\,\mathbf{I}_2\\
            \mathbf{0}_{2\times2} & \mathbf{I}_2
        \end{bmatrix},\quad
        \mathbf{B} = 
        \begin{bmatrix}
            T_s^2/2\,\mathbf{I}_2\\
            T_s\,\mathbf{I}_2
        \end{bmatrix},
    \end{aligned}
\end{equation}
where $T_s$ denotes the sampling period. In addition, the auxiliary matrix 
$\mathbf{C}_{\Dot{\bm{\gamma}}} = [\mathbf{0}_{2\times2}\,\,\mathbf{I}_2]$ extracts the velocity 
from the state, and $v_{\text{max}}$ and $a_{\text{max}}$ denote, respectively, the maximum 
velocity and acceleration that the vehicle may achieve.

The simulation results presented in this section were obtained in MATLAB using the CasADi 
\cite{andersson2012casadi} optimization modeling toolbox, along the IPOPT 
\cite{wachter2006implementation} solver. At each sampling time, the solution obtained at the 
previous step was used to set the initial guess for the current step by performing the shifting 
warm-start method \cite{gros2020linear}. All computations were executed on a single 
desktop computer with an Intel Core i7-6700K @ 4.00 GHz processor and 32.00 GB of RAM.

\subsection{Illustrative Examples} \label{Subsec:IllustrativeExamples}

We begin by presenting some illustrative examples to showcase the trajectories that the algorithm 
is able to produce for different uncertainty maps. In such examples, the drone starts at 
$\mathbf{p} = [1\,\, 1]^\top$ [m] with no initial velocity, and the radius of observation is 
$r = 1$ m. The sampling period is $T_s = 0.1$ s, the horizon is $N = 15$, and the vehicle has a max 
velocity of 4 m/s and a max acceleration of 4 m/$\text{s}^2$. 

\begin{figure}[h]
    \centering 
    \subfloat[Trajectory]{
        \includegraphics[width=0.48\linewidth]{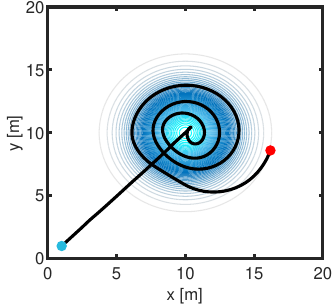}}\hfill
    \subfloat[Sensor footprint]{
        \includegraphics[width=0.475\linewidth]{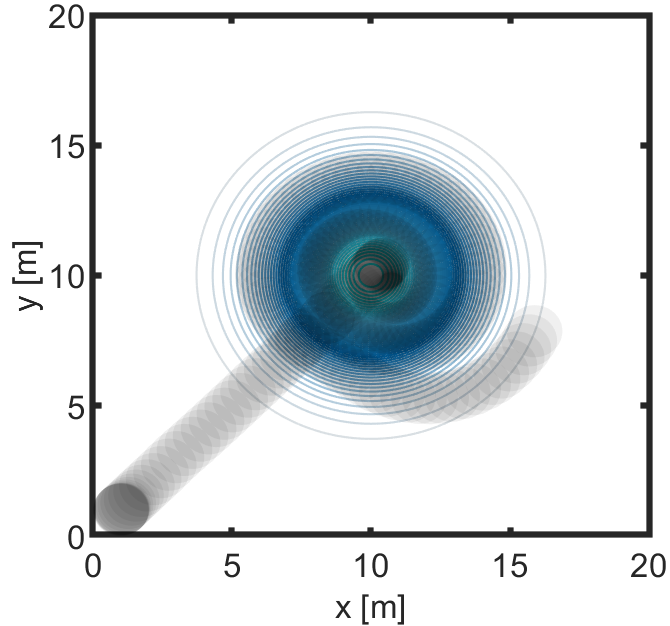}}\\
    \vspace{-1mm}
    \subfloat[Position]{
        \includegraphics[width=0.48\linewidth]{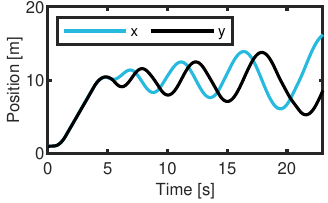}}\hfill
    \subfloat[Uncertainty reduction]{
        \includegraphics[width=0.48\linewidth]{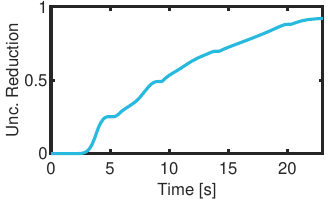}}\\
    \vspace{-1mm}
    \subfloat[Velocity]{
        \includegraphics[width=0.48\linewidth]{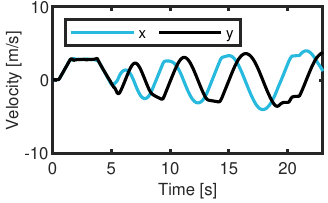}}\hfill
    \subfloat[Computation times]{
        \includegraphics[width=0.48\linewidth]{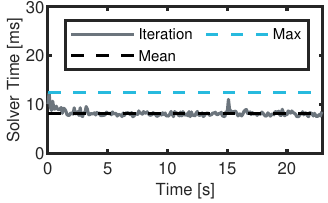}}
    \caption{Example with one radially-symmetric component.}
    \label{Fig:Example1}
\end{figure}

In the first example, illustrated in Fig. \ref{Fig:Example1}, the uncertainty map is composed 
of a single radially-symmetric component. As shown in Fig. \ref{Fig:Example1} (a), initially the 
vehicle moves towards the maximizer of the Gaussian component. Subsequently, as a result of the 
penalties applied by the algorithm, the vehicle moves to wider regions by executing a spiral curve 
with the temporal profiles depicted in Figs. \ref{Fig:Example1} (c) and \ref{Fig:Example1} 
(e). Also, Fig. \ref{Fig:Example1} (b) illustrates the sensor footprint of the 
\gls{UAV}, and Fig. \ref{Fig:Example1} (d) shows the accumulation of the uncertainty volume 
covered by the vehicle over time. In addition, we draw attention to Fig. \ref{Fig:Example1} (f), 
which presents the mean solver times acquired through 100 simulations, with each iteration taking 
approximately 8 ms on average. 

\newpage

\begin{figure}[h]
    \centering 
    \subfloat[Trajectory]{
        \includegraphics[width=0.48\linewidth]{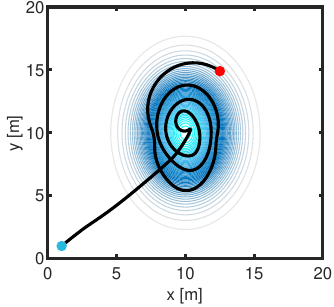}}\hfill
    \subfloat[Sensor footprint]{
        \includegraphics[width=0.475\linewidth]{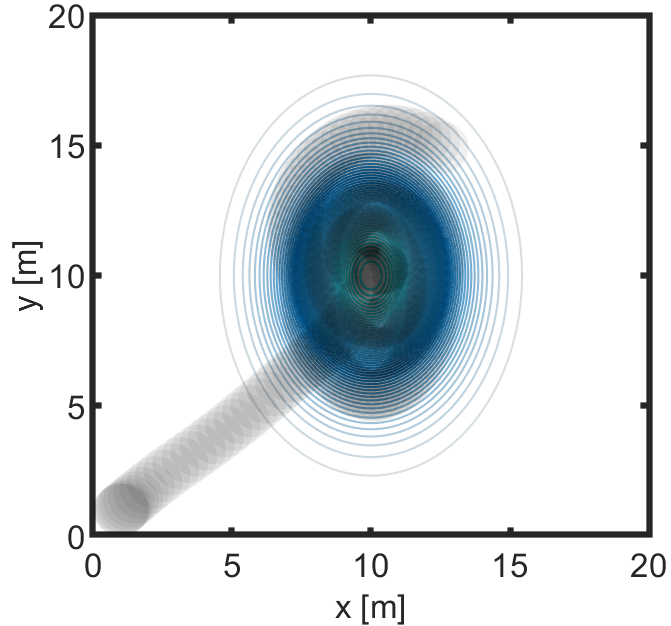}}\\
    \subfloat[Position]{
        \includegraphics[width=0.48\linewidth]{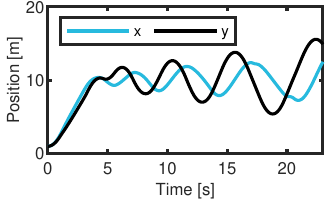}}\hfill
    \subfloat[Uncertainty reduction]{
        \includegraphics[width=0.48\linewidth]{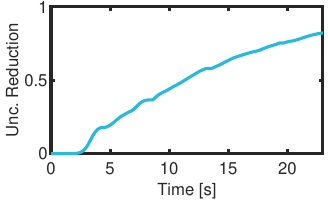}}
    \caption{Example with one elliptical component.}
    \label{Fig:Example2}
\end{figure}

In Fig. \ref{Fig:Example2}, we present another simple example in which the uncertainty map 
consists of a single Gaussian component but now with an elliptical shape. In this case, the 
trajectory adjusts itself to the shape of the component, as can be observed in 
Fig. \ref{Fig:Example2} (a). Furthermore, as shown in Figs. \ref{Fig:Example2} (c) and 
\ref{Fig:Example2} (d), the resulting position and uncertainty reduction profiles are similar to 
those from the previous example.

\newpage

\begin{figure}[h]
    \centering 
    \subfloat[Trajectory]{
        \includegraphics[width=0.48\linewidth]{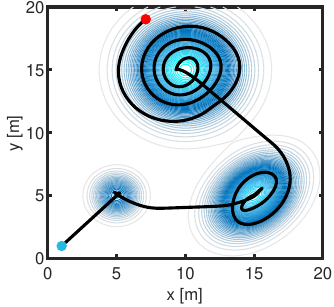}}\hfill
    \subfloat[Sensor footprint]{
        \includegraphics[width=0.475\linewidth]{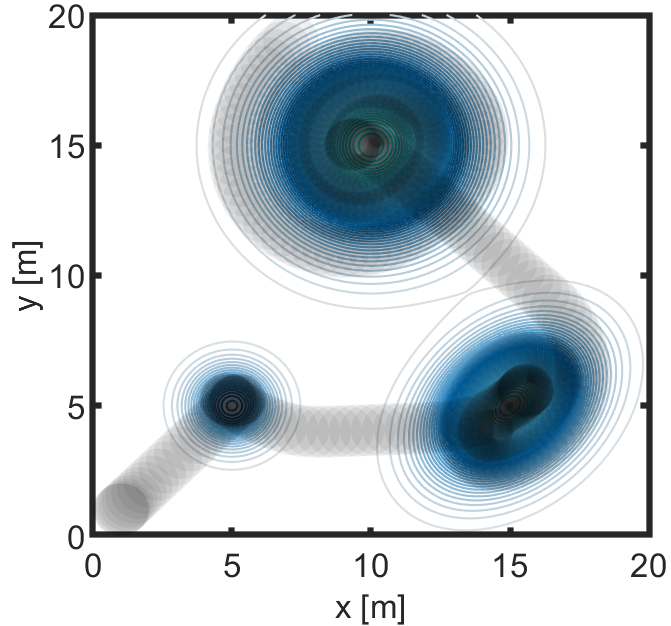}}\\
    \subfloat[Position]{
        \includegraphics[width=0.48\linewidth]{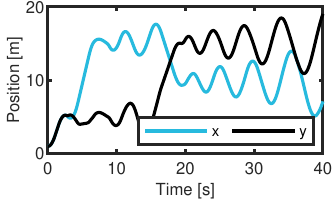}}\hfill
    \subfloat[Uncertainty reduction]{
        \includegraphics[width=0.48\linewidth]{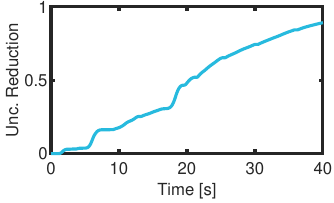}}\\
    \subfloat[Velocity]{
        \includegraphics[width=0.48\linewidth]{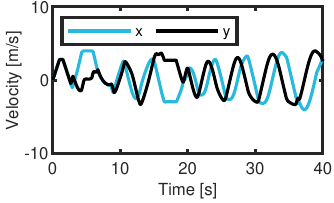}}\hfill
    \subfloat[Computation times]{
        \includegraphics[width=0.48\linewidth]{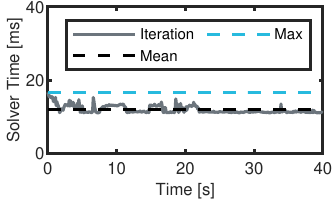}}
    \caption{Example with three Gaussian components.}
    \label{Fig:Example3}
\end{figure}

Now, we consider a more complex example in which the uncertainty function consists of three 
Gaussian components, with the corresponding results displayed in Fig. \ref{Fig:Example3}. As 
depicted in Fig. \ref{Fig:Example3} (a), the drone analyzes each component individually. 
Notably, the components with means at positions $\mathbf{p} = [15\,\, 5]^\top$ and 
$\mathbf{p} = [10\,\, 15]^\top$ exhibit similarities to those in the previous examples, and the 
observed trajectories align with the previous patterns. However, the third component located at 
$\mathbf{p} = [5\,\, 5]^\top$ has a smaller variance when compared to the observation radius of the 
\gls{UAV}.  Consequently, when the drone analyzes this component, it simply hovers at the 
component's maximizer. Additionally, we highlight that the computation times are slightly higher 
in this example, with each iteration averaging approximately 12 ms.

\begin{figure}[h]
    \centering 
    \subfloat[Trajectory]{
        \includegraphics[width=0.48\linewidth]{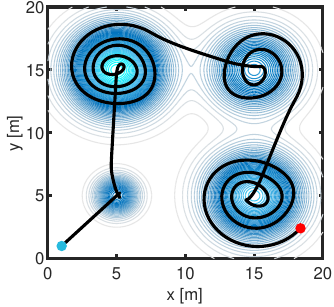}}\hfill
    \subfloat[Sensor footprint]{
        \includegraphics[width=0.475\linewidth]{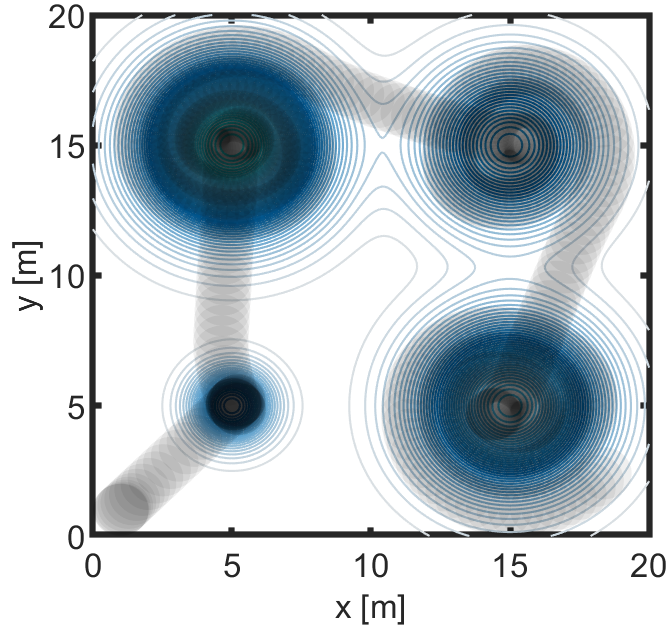}}\\
    \subfloat[Position]{
        \includegraphics[width=0.48\linewidth]{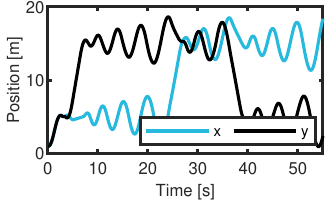}}\hfill
    \subfloat[Uncertainty reduction]{
        \includegraphics[width=0.48\linewidth]{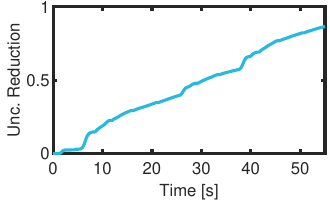}}\\
    \subfloat[Velocity]{
        \includegraphics[width=0.48\linewidth]{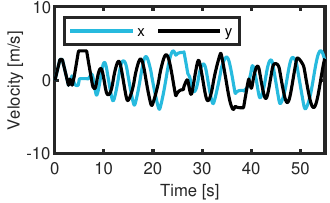}}\hfill
    \subfloat[Computation times]{
        \includegraphics[width=0.48\linewidth]{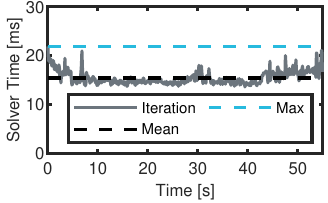}}
    \caption{Example with four radially-symmetric components.}
    \label{Fig:Example4}
\end{figure}

Finally, we introduce an example where the uncertainty map is composed of four radially-symmetric 
Gaussian components, with the corresponding simulation results displayed in Fig. 
\ref{Fig:Example4}. As can be observed in Fig. \ref{Fig:Example4} (a), the component with mean at 
$\mathbf{p} = [5\,\, 5]^\top$ is similar to the one from the previous example, and the remaining 
components all have similar covariance matrices but different associated weights. By observing 
Figs. \ref{Fig:Example4} (a) and \ref{Fig:Example4} (b), one can notice that, as the weights of the 
components increase, the spiral curves become more tightly concentrated, and there is a greater 
overlap of the vehicle's observation circles. In addition, we highlight that, in this example, the 
computational times are slightly higher, with each solver iteration taking approximately 16 ms on 
average, as shown in Fig. \ref{Fig:Example4} (f).

\subsection{Effect of the Weights}

\begin{figure}[h]
    \centering 
    \subfloat[$\lambda = 1/7000$, $\alpha = 0.4$]{
        \includegraphics[width=0.48\linewidth]{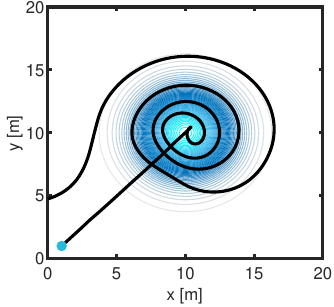}}\hfill
    \subfloat[$\lambda = 1/10000$, $\alpha = 0.4$]{
        \includegraphics[width=0.48\linewidth]{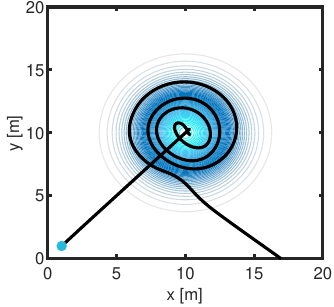}}\\
    \subfloat[$\lambda = 1/7000$, $\alpha = 1$]{
        \includegraphics[width=0.48\linewidth]{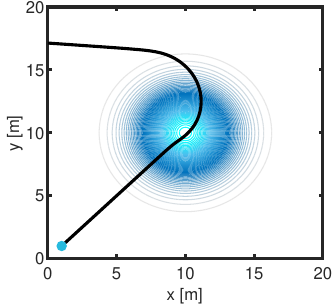}}\hfill
    \subfloat[$\lambda = 1/7000$, $\alpha = 0.8$]{
        \includegraphics[width=0.48\linewidth]{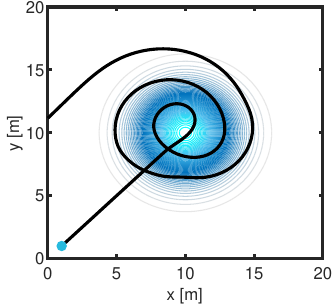}}\\
    \subfloat[Coverage profiles]{
        \includegraphics[width=0.98\linewidth]{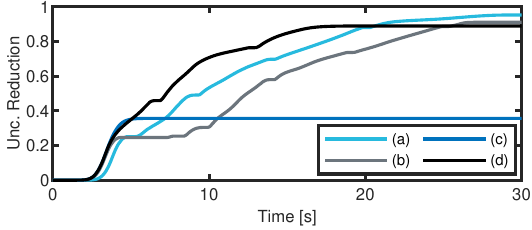}}
    \caption{Results for different combinations of $\lambda$ and $\alpha$.}
    \label{Fig:WeightsEffect}
\end{figure}

As the objective function relies on the exponent $\alpha$ and the scaling coefficient $\lambda$, it 
is worth assessing how these parameters influence the algorithm. In this context, we consider the 
conditions of the initial example, where the uncertainty map comprises a single radially-symmetric 
Gaussian, and we manipulate $\lambda$ and $\alpha$. Fig. \ref{Fig:WeightsEffect} displays the 
trajectories and coverage profiles obtained for some values of $\lambda$ and $\alpha$.

As $\lambda$ decreases in value, less emphasis is placed on the penalty term. Thus, the 
trajectories are expected to become more tightly concentrated, resulting in a greater overlap of 
the observation circles. This effect is noticeable in the examples depicted in 
Figs.\ref{Fig:WeightsEffect} (a) and \ref{Fig:WeightsEffect} (b), and it becomes more pronounced 
when examining Fig. \ref{Fig:WeightsEffect} (e). As shown in Fig. \ref{Fig:WeightsEffect} (e), 
there is a slower initial convergence in the scenario of Fig. \ref{Fig:WeightsEffect} (b) when 
compared to Fig. \ref{Fig:WeightsEffect} (a). However, at $t = 30$ s, both trajectories exhibit a 
similar coverage.

A similar impact can be anticipated when examining the variation of $\alpha$. As $\alpha$ 
increases, the penalization becomes more pronounced, leading to a reduced overlap, as depicted in 
Figs. \ref{Fig:WeightsEffect} (a) and \ref{Fig:WeightsEffect} (d). Particularly, as illustrated in 
Fig. \ref{Fig:WeightsEffect} (e), the trajectory from Fig. \ref{Fig:WeightsEffect} (d) initially 
exhibits a faster convergence than the one from Fig. \ref{Fig:WeightsEffect} (a). However, at 
$t = 30$ s, the trajectory from Fig. \ref{Fig:WeightsEffect} (a) achieves a greater uncertainty 
reduction. Ultimately, Fig. \ref{Fig:WeightsEffect} (c) illustrates a scenario where the value of 
$\alpha$ is sufficiently high to prevent the vehicle from executing a spiral curve.

\newpage

Considering the previous discussion, it becomes clear that there is some need for parameter tuning 
associated with the proposed algorithm. Nevertheless, it should be acknowledged that the algorithm 
has the potential to be extended through the incorporation of variable weights. For instance, one 
could consider assigning higher penalties in regions where the uncertainty is higher, and lower 
penalties in regions where the uncertainty is lower. Moreover, one could employ decaying weights in 
the term $\Tilde{J}_k$ of the objective function to prioritize earlier prediction instants, 
potentially resulting in a faster convergence. Such variations of the algorithm could be easily 
incorporated, and a more exhaustive analysis could be performed. However, the decision to implement 
these variations is left as a user choice and may be a subject of consideration in future research

\subsection{Effect of the Horizon}

\begin{figure}[h]
    \centering 
    \subfloat[$N = 5$]{
        \includegraphics[width=0.48\linewidth]{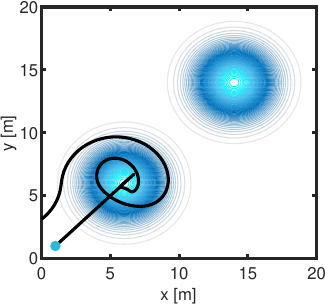}}\hfill
    \subfloat[$N = 15$]{
        \includegraphics[width=0.48\linewidth]{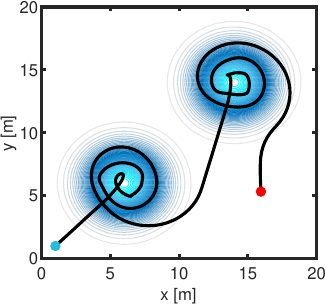}}
    \caption{Resulting trajectories for two horizon lengths.}
    \label{Fig:HorizonEffect}
\end{figure}

It is also important to evaluate how the prediction horizon length impacts the performance of the 
proposed \gls{MPC} algorithm. In this context, we consider* an uncertainty 
map comprising two Gaussian components, with Fig. \ref{Fig:HorizonEffect} depicting the 
generated trajectories for two horizon lengths.

As illustrated in Fig. \ref{Fig:HorizonEffect} (a), for a prediction horizon length of $N = 5$, the 
vehicle's predictive ability falls short and it is not able to predict the second Gaussian 
component. In contrast, when an extended horizon is employed, as shown in 
Fig.\ref{Fig:HorizonEffect} (b) for $N = 15$, the vehicle is able to predict the second Gaussian 
component, leading to a trajectory that covers a greater volume of the uncertainty map.


\section{Experimental Validation} \label{Sec:ExperimentalValidation}

\begin{figure}[b] 
    \centering 
    \includegraphics[width=\linewidth]{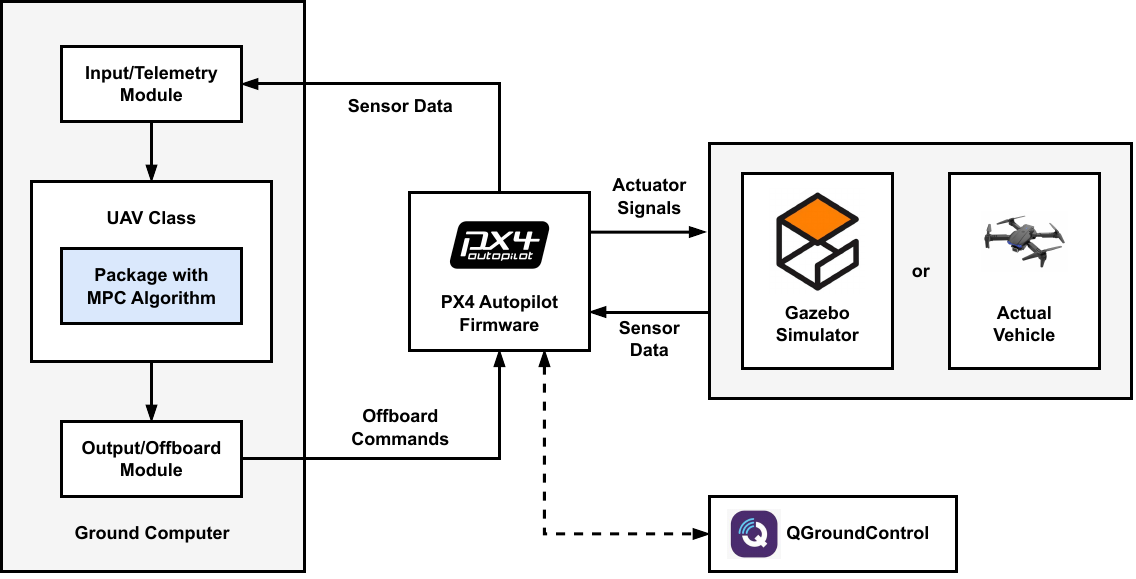} 
    \caption{Employed software architecture.} 
    \label{Fig:Software}
\end{figure}

\begin{figure}[t]
    \centering 
    \subfloat[Iris drone (Gazebo)]{
        \includegraphics[width=0.35\linewidth]{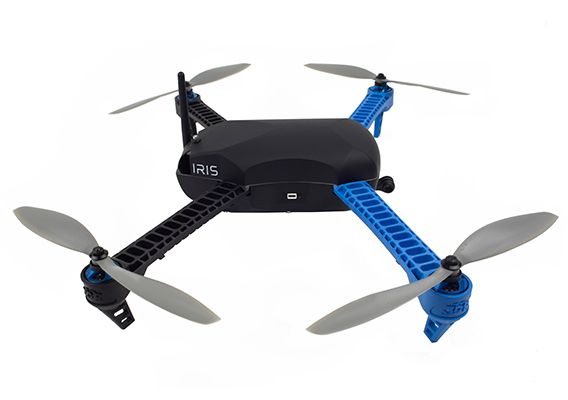}}\hfill
    \subfloat[M690B (field trials)]{
        \includegraphics[width=0.5\linewidth]{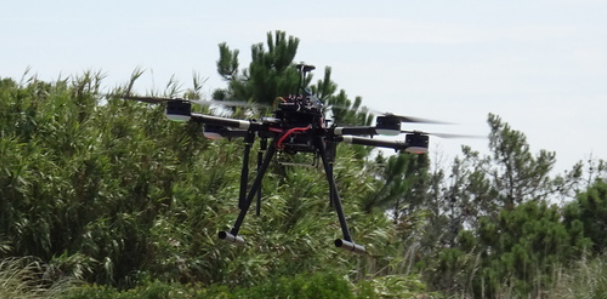}}
    \caption{Quadrotors used in the experimental validation.}
    \label{Fig:Drones}
\end{figure}

In this section, the efficacy of the proposed \gls{MPC} algorithm is assessed through simulations 
in the high-fidelity simulator Gazebo and by conducting actual experiments in an outdoor 
setting. The software used to perform simulations and conduct actual experiments in the quadrotor 
follows from the previous work done by Oliveira \cite{oliveira2021rapid} and Jacinto 
\cite{jacinto2022chemical}, as illustrated in Fig. \ref{Fig:Software}. The operating system 
consists of the Ubuntu 20.04 version along with ROS melodic, and the \gls{MPC} algorithm was 
implemented using the C++ CasADi API. Furthermore, the Gazebo simulations were carried out using 
the Iris model, available through the PX4 Autopilot plugin, and the field trials were conducted 
with the M690B drone from a joint effort between the FirePuma and Capture projects 
\cite{fieldtrials}. Fig. \ref{Fig:Drones} depicts the Iris and the M690B quadrotors used for the 
experimental validation.

Regarding the Gazebo simulations, our initial approaches consisted in providing acceleration and, 
subsequently, velocity references to the PX4 low-level controller. Despite our efforts, these 
approaches posed challenges in achieving smooth and stable trajectories consistent with those 
obtained in MATLAB. Nonetheless, when commanding a trajectory generated offline, it yielded the 
expected outcomes, enabling the vehicle to follow the trajectories with minimal error. 

Despite the lack of significant advantages in executing the algorithm online in this particular
scenario, there is a natural desire to enable the real-time execution of the \gls{MPC}
algorithm to accommodate dynamic alterations in the future, like time-varying uncertainty maps or 
obstacle avoidance. To enable the real-time execution of the algorithm and overcome the poor 
results obtained using lower-level references, we opted for a more conservative approach. The 
approach consists in commanding a given slice of the predicted optimal waypoint sequence to the PX4 
controller. With such an approach, the penalty term of the objective function 
is still updated at each sampling time, but the optimization problem is only solved after the 
application of each waypoint sequence. This method ultimately produced results similar to those 
obtained by instructing a trajectory generated offline.

\subsection{Experimental Results}

In the experiments presented in this section, the drone starts at $\mathbf{p} = [1\,\, 1]^\top$ 
with no initial velocity and the observation radius is assumed to be $r = 0.5$ m. In Gazebo, the 
algorithm operates with a sampling period of $T_s = 0.1$ s, a horizon length of $N = 20$, and the 
first 5 predicted optimal waypoints are sent to the PX4 controller. Consequently, the optimization 
problem is solved from 0.5 s to 0.5 s. The \gls{MPC} is warm-started using the shifting method 
but by shifting 5 steps. Moreover, the \gls{MPC} considers a max velocity of 2 m/s and a max 
acceleration of 2 m/s$^2$ for the drone. Regarding the field tests, due to difficulties faced when 
attempting to execute the algorithm onboard, the field trials were carried out by instructing 
waypoints generated offline.

\begin{figure}[t]
    \centering 
    \hspace{2.5mm}
    \subfloat[Gazebo - Trajectory]{
        \includegraphics[width=0.4\linewidth]{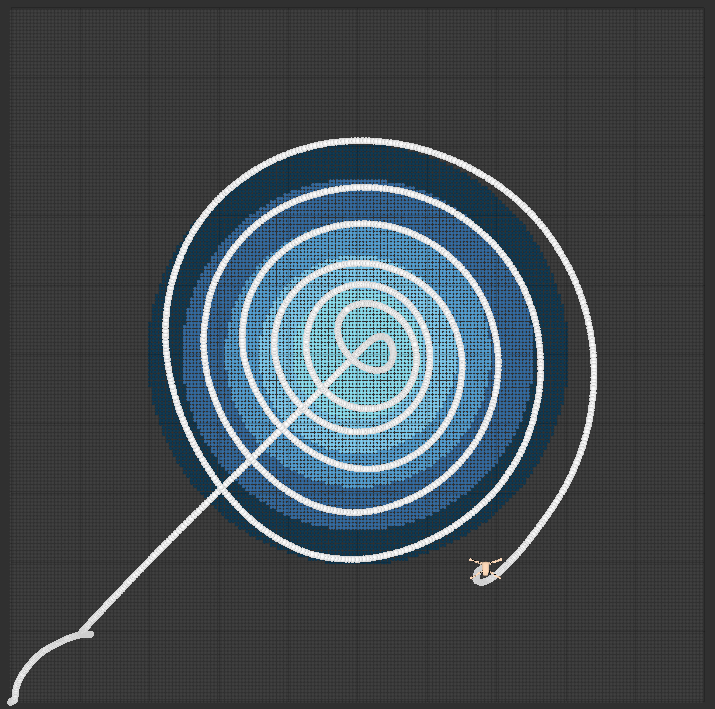}}\hspace{7.5mm}
    \subfloat[Field trial - Trajectory]{
        \includegraphics[width=0.4\linewidth]{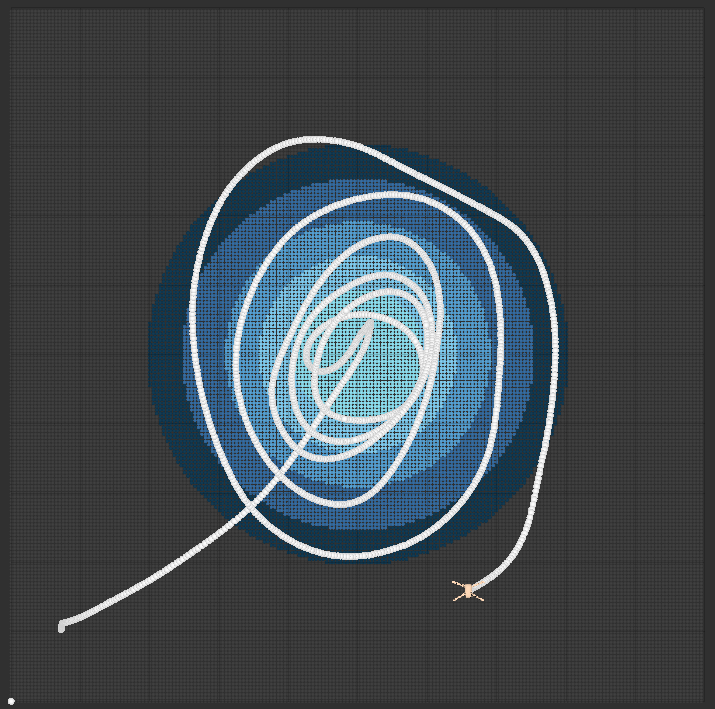}}\\
    \subfloat[Gazebo - Footprint]{
        \includegraphics[width=0.475\linewidth]{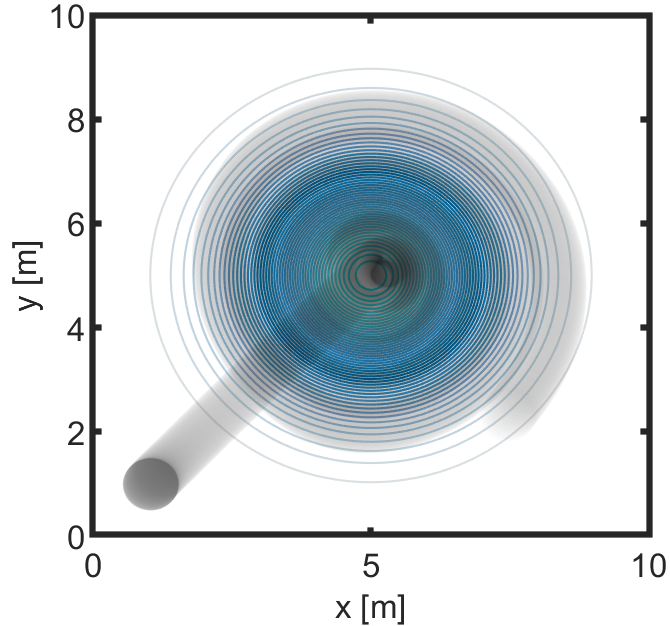}}\hfill
    \subfloat[Field trial - Footprint]{
        \includegraphics[width=0.475\linewidth]{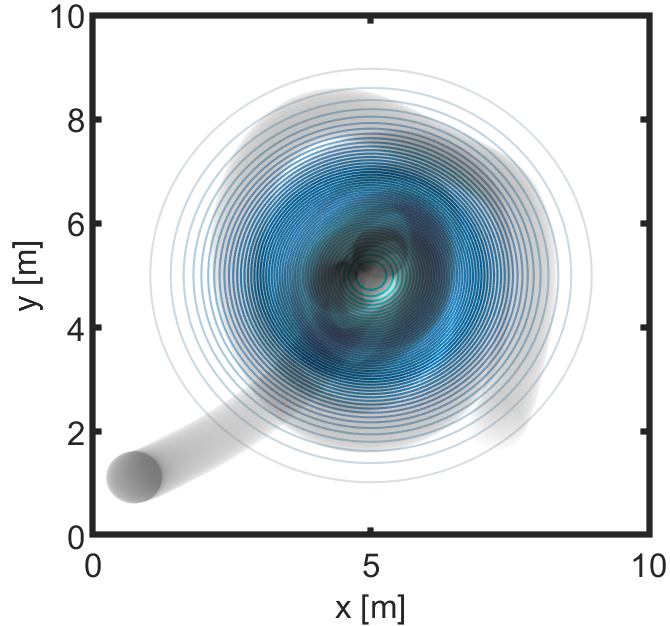}}\\
    \subfloat[Gazebo - Solver]{
        \includegraphics[width=0.48\linewidth]{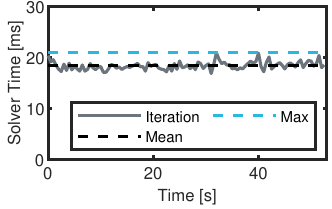}}\hfill
    \subfloat[Uncertainty reduction]{
        \includegraphics[width=0.48\linewidth]{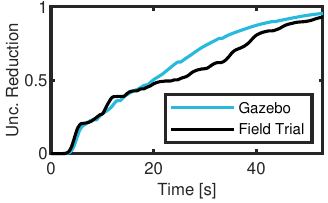}}
    \caption{Gazebo and field trial results for an uncertainty map with a single radially-symmetric component.}
    \label{Fig:Experiment1}
\end{figure}

We begin by considering an example where the uncertainty map is composed of a single 
radially-symmetric component, with the corresponding results displayed in Fig. 
\ref{Fig:Experiment1}. As illustrated in Fig. \ref{Fig:Experiment1} (a), the drone exhibits the 
expected behavior in the Gazebo simulation, executing a smooth spiral curve, as also reflected in 
the position profiles shown in Fig. \ref{Fig:Experiment1} (e). This behavior is obviously possible 
due to the appropriately tuned parameters of the \gls{MPC}, which allow the generation of smooth 
trajectories that the PX4 controller can track efficiently. In addition, as depicted in Fig. 
\ref{Fig:Experiment1} (g), the computational times are also sufficiently fast to allow for a 
good performance, with each solver iteration taking approximately 18 ms on average.

Concerning the field trial, as shown in Fig. \ref{Fig:Experiment1} (b), it can be observed 
that the resulting trajectory is not as consistent as the one from the Gazebo simulation. This 
discrepancy primarily arises from the influence of wind disturbances encountered in the outdoor 
experimental environment. In addition, there are also some inaccuracies associated with the 
\gls{GPS} of the drone, structural differences between the drones used in Gazebo and in the 
real trials, and differences in the tuning of the PX4 inner-loop controllers. Nevertheless, for the 
designated observation radius, both trajectories show a similar coverage by the final instant, 
as shown in Fig. \ref{Fig:Experiment1} (h).

\begin{figure}[t]
    \centering 
    \hspace{2.5mm}
    \subfloat[Gazebo - Trajectory]{
        \includegraphics[width=0.4\linewidth]{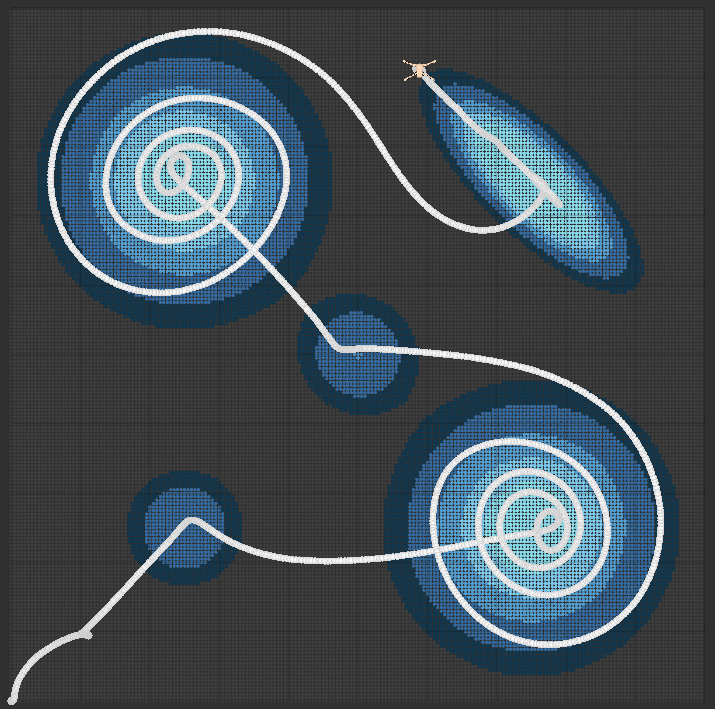}}\hspace{7.5mm}
    \subfloat[Field trial - Trajectory]{
        \includegraphics[width=0.4\linewidth]{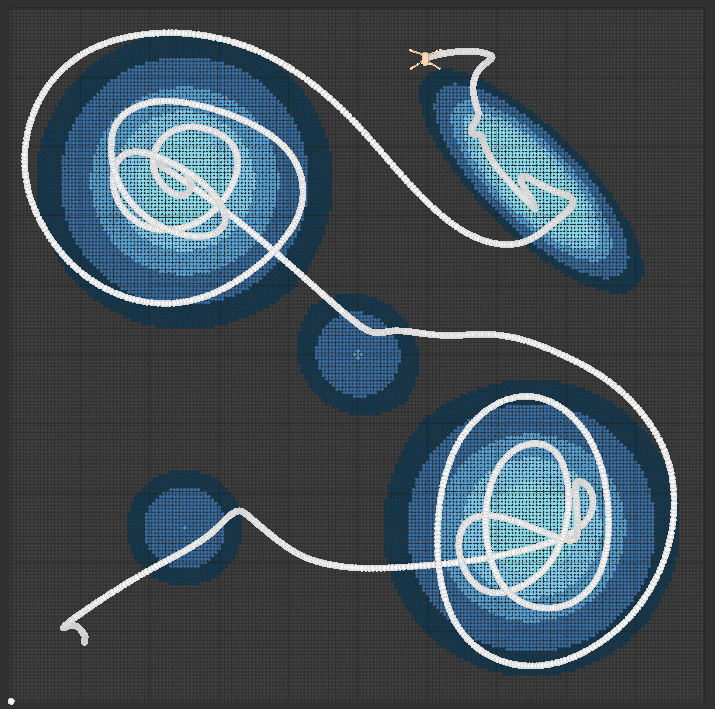}}\\
    \subfloat[Gazebo - Footprint]{
        \includegraphics[width=0.475\linewidth]{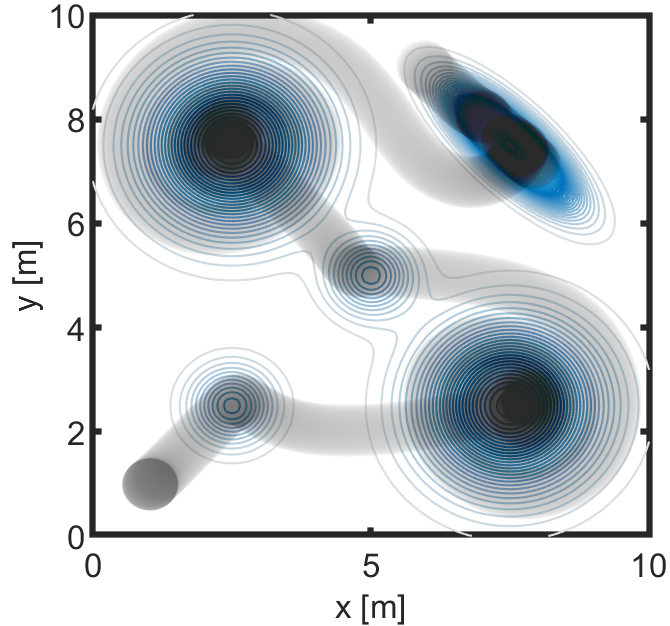}}\hfill
    \subfloat[Field trial - Footprint]{
        \includegraphics[width=0.475\linewidth]{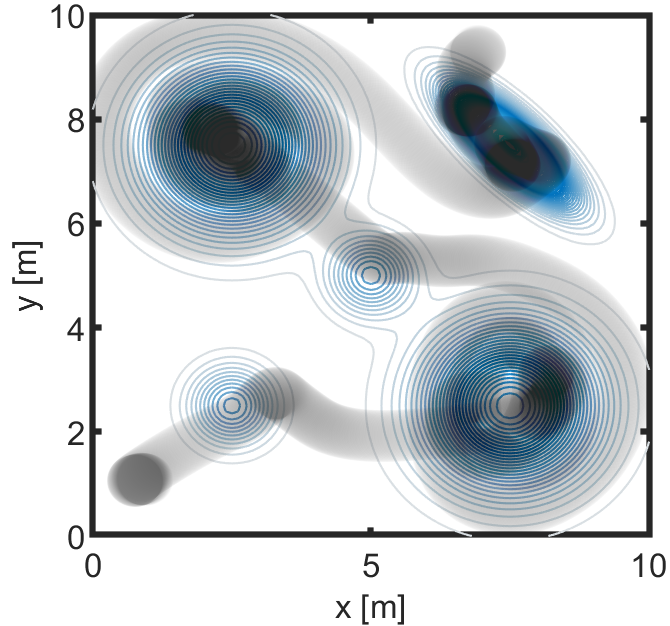}}\\
    \subfloat[Gazebo - Solver]{
        \includegraphics[width=0.48\linewidth]{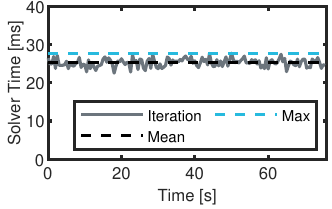}}\hfill
    \subfloat[Uncertainty reduction]{
        \includegraphics[width=0.48\linewidth]{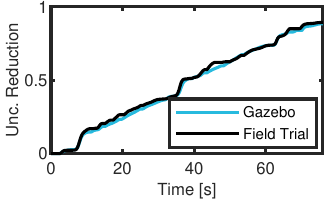}}
    \caption{Gazebo and field trial results for an uncertainty map comprising five Gaussian components.}
    \label{Fig:Experiment2}
\end{figure}

To conclude, we present an example where the uncertainty map comprises five Gaussian components, 
with Fig. \ref{Fig:Experiment2} displaying the results obtained in Gazebo and in the 
corresponding field trial. As illustrated in Fig. \ref{Fig:Experiment2} (a), the 
map comprises four circular Gaussian components. Two of these components have relatively small 
variances in comparison to the vehicle's observation radius, while the other two exhibit higher 
variances. Additionally, there is a fifth component with an elliptical shape, and its variance 
along one of its axes is small when compared to the radius of observation.

In particular, we draw attention to the vehicles's behavior when analyzing the fifth component, in 
which case the drone follows a straight path along the major axis of the Gaussian. In addition, we 
draw attention to Fig. \ref{Fig:Experiment2} (g), which shows that each solver iteration now 
takes approximately 25 ms on average. Ultimately, we highlight that the field trial results are 
comparable to the Gazebo simulation.

\section{Conclusion} \label{Sec:Conclusion} 

This paper tackles the trajectory planning problem for \gls{UAV} search and coverage missions 
based on an uncertainty map described as a linear combination of Gaussian distributions. We propose 
an \gls{MPC} algorithm that promotes the exploration of the map by preventing the vehicle from 
revisiting previously covered regions. This is achieved by penalizing intersections between 
the circular observation regions along the vehicle's trajectory. Due to the complexity of precisely 
determining the intersection area between two circles, we introduce an exponential penalty 
function. The algorithm is tested in MATLAB, Gazebo, and in outdoor trials. The results show 
that the algorithm can generate efficient trajectories for search and coverage missions. 

Possible extensions involve developing a subroutine to reduce the number of components in the 
penalty term and generalizing the algorithm using variable weights to finetune its performance. 
Since we assumed a static uncertainty map, future research may also focus on the search and 
coverage problem based on time-varying uncertainty functions.


\bibliographystyle{ieeetr}

\bibliography{Refs}

\bio{Figs/HugoMatias}
Hugo Matias received his B.Sc. and M.Sc. degrees in Electrical and Computer Engineering from the 
Instituto Superior Técnico (IST), Lisbon, Portugal, in 2021 and 2023, respectively. His main area 
of specialization is Control, Robotics and Artificial Intelligence, and his second area of 
specialization is Networks and Communication Systems. He is pursuing a Ph.D. in Electrical and 
Computer Engineering, with a specialization in Systems, Decision and Control, at the School of 
Science and Technology from the NOVA University of Lisbon, Costa da Caparica, Portugal, and he is 
also conducting his research at the Institute for Systems and Robotics (ISR), LARSyS, Lisbon, 
Portugal. His research interests span the fields of nonlinear control and optimization, filtering 
and estimation, and distributed systems.
\endbio

\bio{Figs/Daniel}
Daniel Silvestre received his B.Sc. in Computer Networks in 2008 from the Instituto 
Superior Técnico (IST), Lisbon, Portugal, and his M.Sc. in Advanced Computing in 2009 from the 
Imperial College London, London, United Kingdom. In 2017, he received his Ph.D. (with the highest 
honors) in Electrical and Computer Engineering from the former university. Currently, he is with 
the School of Science and Technology from the Nova University of Lisbon and also with the Institute 
for Systems and Robotics at the Instituto Superior Técnico in Lisbon (PT). His research interests 
span the fields of fault detection and isolation, distributed systems, guaranteed state estimation, 
computer networks, optimal control and nonlinear optimization.
\endbio

\end{document}